\documentclass[letterpaper]{article} 
\usepackage{aaai2026}
\usepackage{booktabs} 
\usepackage{multirow} 
\usepackage{times}  
\usepackage{helvet}  
\usepackage{courier}  
\usepackage[hyphens]{url}  
\usepackage{graphicx} 
\usepackage{natbib}  
\usepackage{caption} 
\usepackage{algorithm}
\usepackage{algorithmic}

\usepackage{amsmath}
\usepackage{pdfpages}
\usepackage{tabularx}
\usepackage{multirow}
\usepackage{booktabs}
\usepackage{adjustbox}

\usepackage{newfloat}
\usepackage{listings}
\DeclareCaptionStyle{ruled}{labelfont=normalfont,labelsep=colon,strut=off} 
\floatstyle{ruled}
\newfloat{listing}{tb}{lst}{}
\floatname{listing}{Listing}
\title{AgentCDM: Enhancing Multi-Agent Collaborative Decision-Making via ACH-Inspired Structured Reasoning}

\author {
    Xuyang Zhao\textsuperscript{\rm 1},
    Shiwan Zhao\textsuperscript{\rm 1},
    Hualong Yu\textsuperscript{\rm 1},
    Liting Zhang\textsuperscript{\rm 1},
    Qicheng Li\textsuperscript{\rm 1}$^{*}$
}
\affiliations {
    \textsuperscript{\rm 1}College of Computer Science, Nankai University, Tianjin, China\\
    xychao@mail.nankai.edu.cn, liqicheng@nankai.edu.cn
}

\usepackage{bibentry}

\begin{document}

\maketitle

\begin{abstract}
Multi-agent systems (MAS) powered by large language models (LLMs) hold significant promise for solving complex decision-making tasks. However, the core process of collaborative decision-making (CDM) within these systems remains underexplored. Existing approaches often rely on either ``dictatorial" strategies that are vulnerable to the cognitive biases of a single agent, or ``voting-based" methods that fail to fully harness collective intelligence. To address these limitations, we propose \textbf{AgentCDM}, a structured framework for enhancing collaborative decision-making in LLM-based multi-agent systems. Drawing inspiration from the Analysis of Competing Hypotheses (ACH) in cognitive science, AgentCDM introduces a structured reasoning paradigm that systematically mitigates cognitive biases and shifts decision-making from passive answer selection to active hypothesis evaluation and construction. To internalize this reasoning process, we develop a two-stage training paradigm: the first stage uses explicit ACH-inspired scaffolding to guide the model through structured reasoning, while the second stage progressively removes this scaffolding to encourage autonomous generalization. Experiments on multiple benchmark datasets demonstrate that AgentCDM achieves state-of-the-art performance and exhibits strong generalization, validating its effectiveness in improving the quality and robustness of collaborative decisions in MAS.

\end{abstract}

\section{Introduction}
Large Language Models (LLMs) have achieved remarkable success across a wide range of natural language processing tasks, owing to their strong capabilities in language understanding and reasoning~\cite{achiam2023gpt,guo2025deepseek,grattafiori2024llama}. However, despite their impressive performance, individual LLMs still exhibit several fundamental limitations, such as potential security vulnerabilities~\cite{wolf2023fundamental}, content hallucination~\cite{min2023factscore}, and poor handling of complex, multi-step reasoning tasks~\cite{hadi2023large}. These limitations have motivated increasing interest in developing multi-agent systems (MAS) built on top of LLMs, aiming to leverage agent collaboration to overcome the deficiencies of a single model.

In LLM-based MAS, individual agents are typically assigned specialized roles~\cite{li2023camel}, and through interaction and collaboration, they can collectively solve complex problems that are challenging for a single agent to address alone. This paradigm has demonstrated promising potential in various domains, including code generation~\cite{huang2023agentcoder}, web browsing~\cite{chen2024mindsearch}, scientific exploration~\cite{lu2024ai}, and tool use~\cite{shen2024small}. Over the past year, the field has evolved rapidly—from static, hand-crafted frameworks to more dynamic architectures in which agent roles and behaviors are adaptive and context-aware. This evolution suggests a clear trajectory toward higher levels of automation and generalization in multi-agent collaboration.

Despite this progress, most existing research on LLM-based MAS has focused primarily on designing interaction protocols—such as communication, coordination, or planning—while largely overlooking the core process of collaborative decision-making (CDM). As the component that ultimately determines the system's output, CDM is critical to the quality and robustness of MAS. Yet prevailing decision mechanisms are often simplistic, typically falling into one of two categories: ``dictatorial" or ``voting-based". In dictatorial schemes, a single agent serves as the final decision-maker, rendering the outcome highly vulnerable to that agent's internal limitations, such as role confusion, instruction repetition, or infinite loops~\cite{li2023camel}. In voting-based methods, while multiple agents contribute, the aggregation process is typically shallow, failing to synthesize partial truths or resolve contradictions. 

At the heart of these limitations lies a deeper problem: \textbf{cognitive bias}. Biases such as anchoring, confirmation bias, and subjective validation affect both individual agents and collective decision-making processes. In dictatorial CDM, the final output is shaped by the subjective inclinations of a single agent. In voting-based CDM, aggregated outputs often retain unchallenged or inconsistent reasoning patterns. These biases degrade the system's ability to reason thoroughly and reliably, especially in high-stakes or ambiguous tasks. Thus, addressing cognitive bias is not only desirable but necessary to advance the effectiveness of CDM in LLM-based MAS.

\begin{figure*}
    \centering
    \includegraphics[width=1\linewidth,height=9cm]{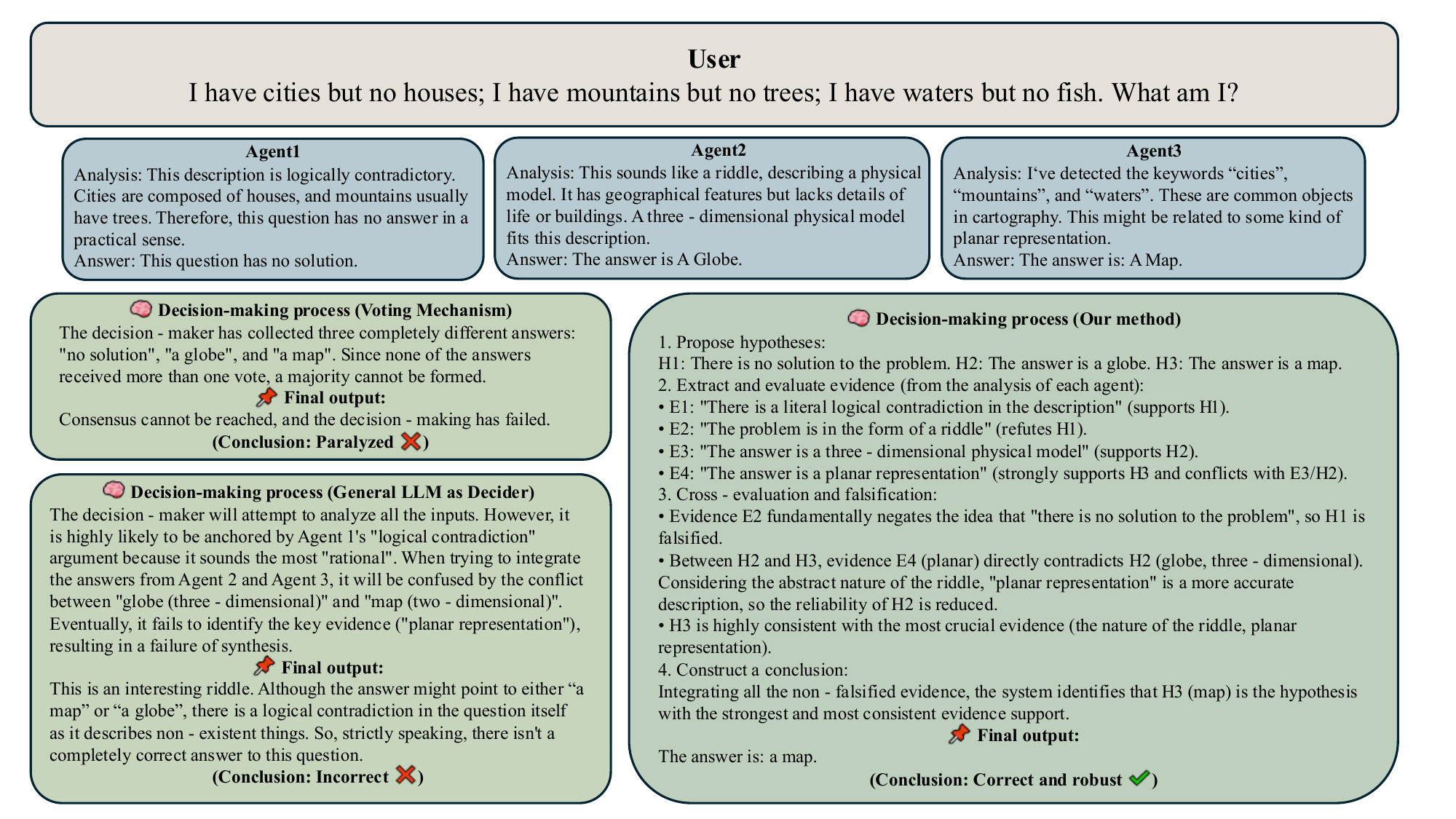}
    \caption{Comparison of collaborative decision-making strategies: (a) Voting-based methods fails due to a lack of consensus among agents, resulting in decision paralysis. (b) Dictatorial strategies rely on a single agent's judgment, which may be biased or unstable. (c) Our ACH-inspired protocol guides the decision agent through structured hypothesis evaluation, enabling more robust and bias-resistant reasoning.}
    \label{fig1}
\end{figure*}

In human decision-making, structured analytical frameworks are widely used to mitigate cognitive bias. One such framework is the \textit{Analysis of Competing Hypotheses} (ACH)~\cite{heuer1999psychology}, a systematic method developed in cognitive science to help analysts evaluate multiple competing explanations based on evidence, promoting disconfirmation and deliberative reasoning. Inspired by ACH, we incorporate a structured reasoning strategy into our decision-making agent, guiding it to evaluate alternative hypotheses and synthesize a superior answer through evidence-driven analysis. A conceptual comparison of this ACH-based strategy against existing methods is provided in Figure~\ref{fig1}.

To internalize this structured reasoning process within the model, we introduce \textbf{AgentCDM}, a novel two-stage training framework. In the first stage, agents are trained with explicit ACH-inspired scaffolding that demonstrates how to conduct structured, bias-resistant reasoning. In the second stage, this scaffolding is gradually removed, encouraging agents to generalize the learned reasoning strategies and apply them autonomously in novel contexts. This design enables the agent to go beyond answer selection—toward the construction of higher-quality, collectively informed decisions.

\vspace{0.5em}
\noindent\textbf{Our contributions are summarized as follows:}
\begin{itemize}
    \item We introduce a structured reasoning strategy inspired by cognitive science (ACH) to systematically mitigate cognitive bias in LLM-based collaborative decision-making.
    \item We propose \textbf{AgentCDM}, a two-stage training paradigm that enables decision-making agents in MAS to internalize and generalize ACH-inspired reasoning processes.
    \item We conduct extensive experiments on multiple benchmark datasets, demonstrating that AgentCDM achieves state-of-the-art performance and exhibits strong generalization across domains.
\end{itemize}

\section{Related Work}

\subsection{LLM-based Multi-Agent Systems}

With the advancement of LLMs, multi-agent systems (MAS) leveraging LLMs have become a significant approach to augmenting task-processing capabilities. Frameworks such as CAMEL~\cite{li2023camel}, MetaGPT~\cite{hong2023metagpt}, and AutoGen~\cite{wu2024autogen} enable agents to interact and coordinate through role-playing, predefined workflows, or multi-agent conversations. SWIFTSAGE~\cite{lin2023swiftsage} further draws on dual-process cognition to enhance efficiency and robustness.

While extensive attention has been paid to agent interaction protocols, there is relatively less focus on the aggregation methods for integrating agent outputs into cohesive final decisions.

\subsection{Collaborative Decision-Making in MAS}

Collaborative decision-making (CDM) in MAS aims to achieve collectively superior outcomes through agent cooperation~\cite{bose2017collective,king2007use}. Existing LLM-based MAS decision-making broadly falls into two categories: dictatorial and voting-based methods.

Dictatorial approaches appoint a predefined ``decision-maker" (e.g., leader, critic, judge) to synthesize inputs and render final judgments~\cite{hao2025chatllm,li2023camel,liang2023encouraging}. However, these methods heavily depend on the reasoning capacity of the central agent, with limited research on systematically improving its robustness and interpretability, especially under conflicting inputs. This vulnerability is compounded by the fact that these decision-makers are typically activated by unstructured prompts, which simply request a final verdict without providing a systematic framework for resolving contradictions or mitigating cognitive biases.

Voting-based mechanisms aggregate agent outputs through majority or peer review paradigms~\cite{li2024more,du2023improving,xu2023towards}, leveraging collective intelligence for more robust reasoning. Nonetheless, these methods are susceptible to individual agent unreliability and cognitive biases.

Overall, current approaches either centralize decision responsibilities or rely on simple aggregation, both with notable limitations in analytical depth and objectivity.

\subsection{Rule-Based Reinforcement Learning}

The use of reinforcement learning (RL) to enhance LLM reasoning has been notably advanced by DeepSeek-R1, which employs rule-based rewards and critic-free algorithms like Group Relative Policy Optimization (GRPO)~\cite{guo2025deepseek}. Subsequent analyses identified and mitigated GRPO's optimization biases (e.g., Dr.GRPO~\cite{liu2025understanding}), proposed simplifications such as Reinforce-Rej~\cite{xiong2025minimalist} and DAPO~\cite{yu2025dapo}, and ultimately removed policy constraints in GPG for further minimalism~\cite{chu2025gpg}.

These developments highlight a progression from paradigm innovation to analytical refinement in rule-based RL for LLMs. However, the application of this paradigm to multi-agent settings remains largely unexplored.

\section{Problem Definition and Formulation}
We formalize the process of an MAS handling a user query $s$. The system comprises $n$ \textit{Execution Agents} $\{\pi_1, \pi_2, \dots, \pi_n\}$ and a single \textit{Decision Agent} $\pi_D$. The overall workflow is divided into two core phases:

\paragraph{Execution Phase}
In this phase, each agent $\pi_i$ (where $i \in \{1,2,\dots,n\}$) independently receives the user query $s$ and generates a candidate answer $a_i$ according to its own policy $\pi_i(\cdot \mid s)$. Formally, this process can be represented as:
\begin{equation}
  a_i \sim \pi_i(\cdot|s),\forall i \in \{1,2,\dots,n\}.
\end{equation}

\paragraph{Decision Phase}
The Decision Agent $\pi_D$ receives the full context, including the original query $s$ and all candidate answers $\{a_1, a_2, \dots, a_n\}$ from the execution agents. We define this context as the history $\mathbf{H} = \{s, a_1, a_2, \dots, a_n\}$. Based on $\mathbf{H}$, $\pi_D$ performs reasoning and produces the final, unified answer $a_D$ for the system:
\begin{equation}
  a_D \sim \pi_D(\cdot|H).
\end{equation}

\section{Method}
Although LLM-based MAS show great potential for solving complex problems through collective intelligence, how to effectively conduct the final collaborative decision remains an underexplored area. When interacting with a user, the system must provide a single, coherent output, which requires a powerful decision agent to act as the ``final arbiter". Its necessity is twofold: 1) \textbf{Single Point of User Interaction}: The user needs a clear response, not multiple, potentially conflicting or redundant answers. 2) \textbf{Conflict Resolution and Information Synthesis}: When different agents provide contradictory or complementary information, a decision-making mechanism is needed to arbitrate, integrate, and construct the highest-quality answer.

However, training a decision agent with these capabilities presents significant challenges. Traditional methods like SFT struggle to internalize complex and robust reasoning capabilities into the model, and its generalization ability is limited, especially when facing conflicts or inconsistencies among agents.

\begin{figure}[!ht]
    \centering
    \includegraphics[width=1\linewidth]{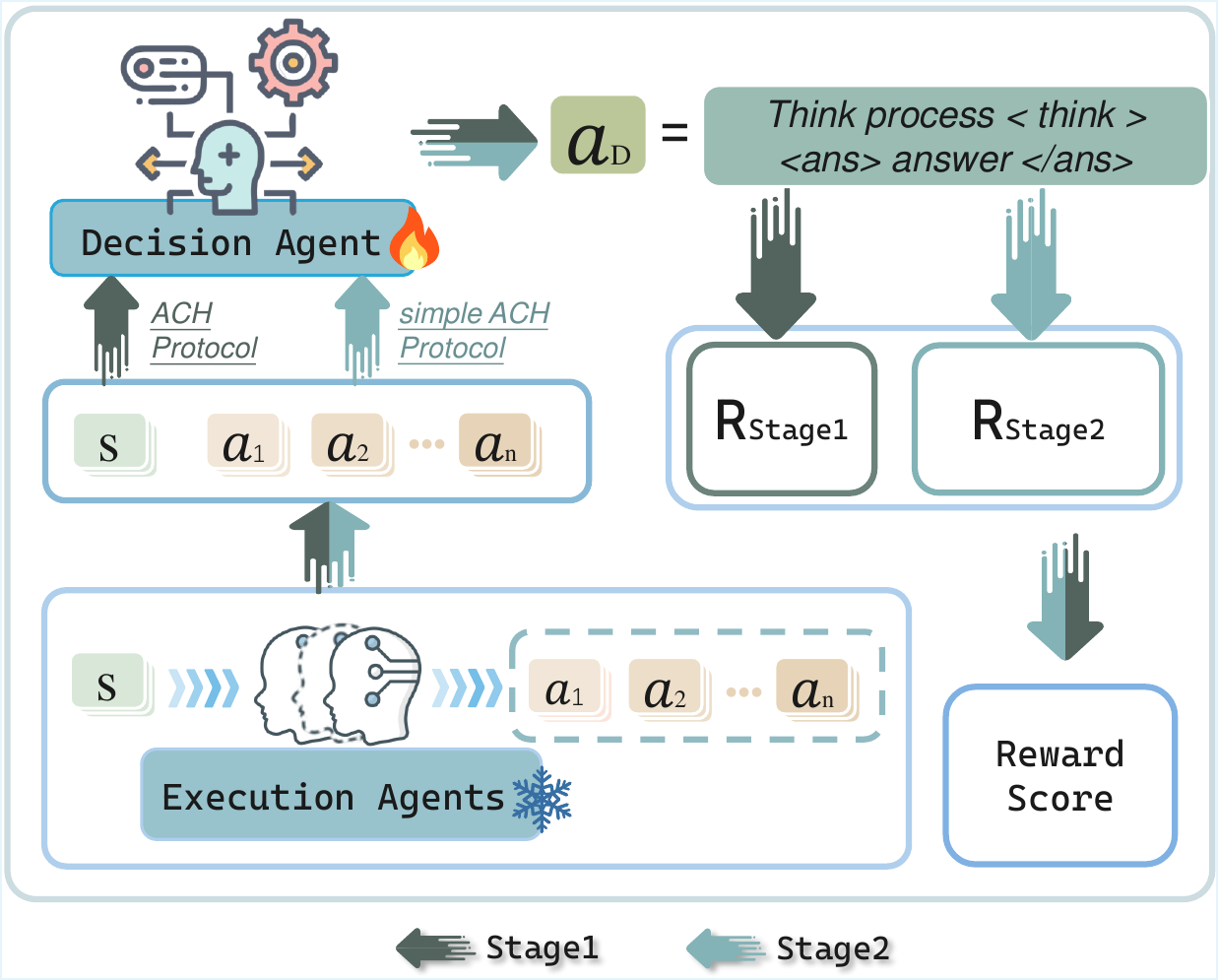}
    \caption{The AgentCDM framework. In the first stage, the framework uses a detailed ACH (Analysis of Competing Hypotheses) prompt as a structural guide or ``scaffold". In the second stage, it employs a curriculum learning strategy to progressively remove this scaffold, encouraging the agent to learn and internalize the reasoning process. }
    \label{fig2}
\end{figure}

To this end, we propose \textbf{AgentCDM}, a training framework designed to enhance an LLM's ability to act as a decision-maker in an MAS. Our goal is to enable the model not only to integrate responses from different agents but, more importantly, to generate a high-quality and robust final decision by applying structured reasoning, especially when faced with conflicts and contradictions. The overall architecture of this framework, illustrated in Figure~\ref{fig2}, consists of two core operational phases (Execution and Decision) and is underpinned by a novel two-stage training paradigm, which we will describe in detail throughout this section.

\subsection{Stage One: Structured Reasoning Protocol based on ACH}

Recent research, such as DeepSeek-R1~\cite{guo2025deepseek}, has demonstrated that reinforcement learning can effectively unlock the latent reasoning capabilities of pretrained language models. Building on this insight, we aim to equip the decision-making agent with more structured and cognitively aligned reasoning abilities early in training. To this end, our framework introduces explicit ``scaffolding" in the first stage: a reasoning protocol inspired by the ACH. This protocol acts as an external guide, helping the agent learn to decompose decision problems, evaluate competing hypotheses, and reach bias-resistant conclusions through systematic analysis. Without such scaffolding, the model struggles to autonomously develop the desired reasoning structures, often reverting to shallow heuristics or inconsistent judgment patterns.

\begin{figure}[!ht]
    \centering
    \includegraphics[width=1\linewidth]{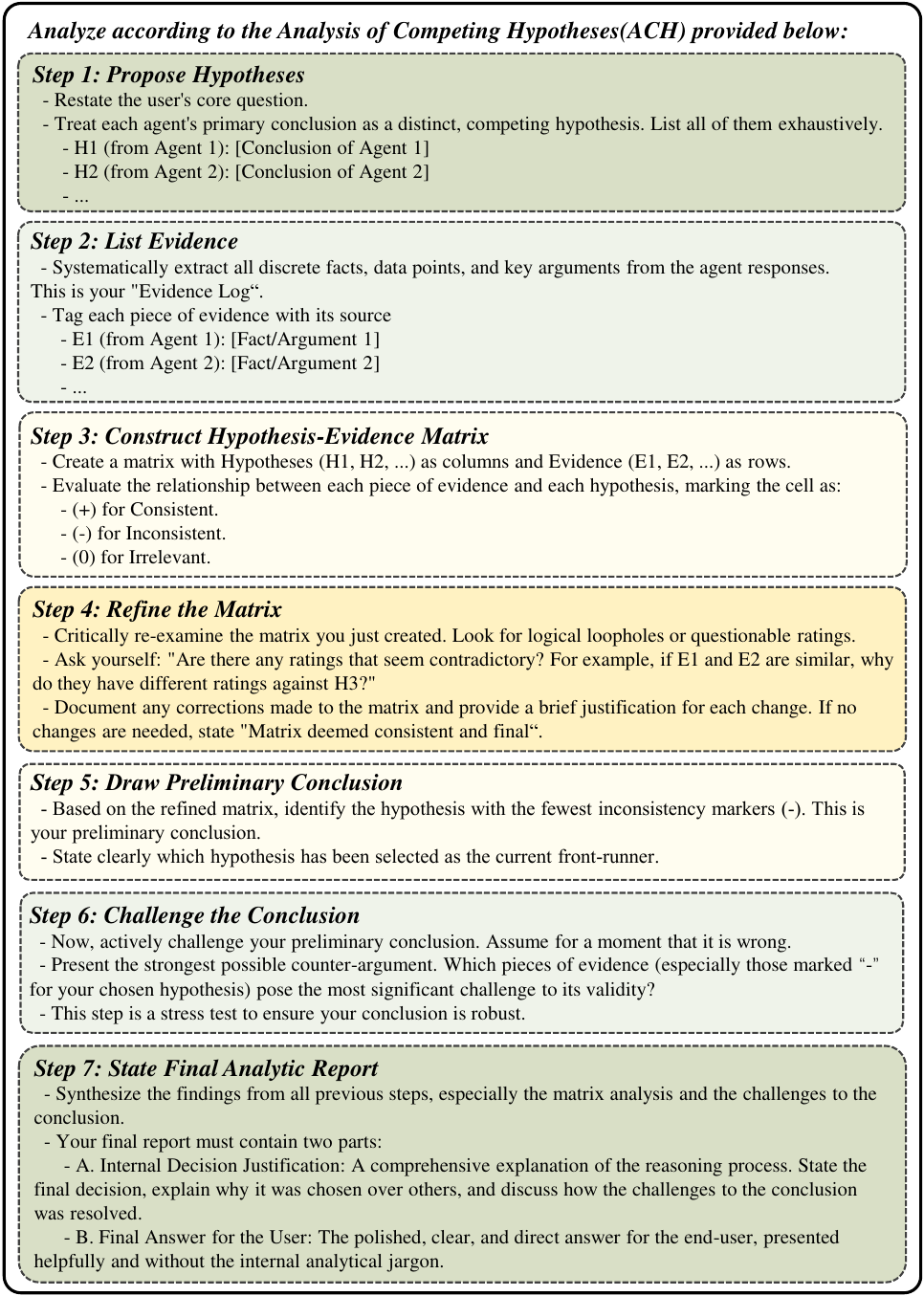}
    \caption{Decision-Making Process Based on Analysis of Competing Hypotheses}
    \label{fig3}
\end{figure}

The ACH Protocol we proposed, illustrated in Figure~\ref{fig3}, begins by formulating a comprehensive and mutually exclusive set of hypotheses to establish an unbiased \textit{hypothesis space}. Concurrently, it requires the systematic collection of all relevant facts and arguments into a shared \textit{evidence pool} before evaluation begins. The core of the framework is a \textit{hypothesis-evidence matrix} that cross-evaluates the diagnostic value of each piece of evidence against every hypothesis (as consistent, inconsistent, or irrelevant). A crucial \textit{meta-cognitive review} step is integrated to scrutinize and correct potential biases in this evaluation, ensuring analytical integrity. In drawing a conclusion, the protocol deliberately shifts the focus from confirmation to \textit{falsification}, preliminarily selecting the hypothesis with the least amount of disconfirming evidence. This tentative conclusion is then immediately subjected to rigorous \textit{adversarial testing} to proactively probe its robustness. Finally, the framework culminates in a comprehensive analytical report that not only articulates the final decision and its key evidentiary support but also details the rationale for rejecting alternative hypotheses and assesses the relative strengths and weaknesses of the conclusion.

In RL, the reward function is the core signal that guides model optimization. To train the decision-maker agent, we have designed a composite, rule-based reward function consisting of three parts:

\paragraph{Format rewards}
We require the model to encapsulate its thinking process within \texttt{<think>} tags and its final answer within \texttt{<answer>} tags. This design not only encourages the model to develop the habit of ``thinking before answering", but also establishes a standardized output format that facilitates accurate downstream parsing and evaluation.

\paragraph{Accuracy rewards}
This reward is used to evaluate the correctness of the final answer given by the model. With the guidance of the format reward, the model provides the answer in the correct format, allowing us to provide feedback on correctness in a rule-based manner.

\paragraph{ACH rewards}
The core objective of this stage is to guide the model to strictly follow the ACH Protocol in its thinking process. We use pattern matching to verify whether the model's output within the \texttt{<think>} tags adheres to the ACH Protocol, thereby supervising its reasoning process.

Thus, the final score is computed as:
\begin{equation}
  score_\text{Stage1} = score_\text{format} + score_\text{answer} + score_\text{ACH}.
\end{equation}

\subsection{Stage Two: Scaffolding Removal and Autonomous Exploration}

The training in the first stage equips the model with the ability to apply the ACH Protocol under explicit guidance. To encourage the model to internalize this capability and to autonomously explore superior decision-making paths without ``scaffolding", we designed a second stage. Directly removing the ACH Protocol would lead to training collapse, so we introduced a smoother transition mechanism.

\paragraph{Soft ACH rewards}
In the second stage of training, we transition from strict pattern matching to a more nuanced guidance mechanism. We introduce a soft structural reward, denoted as $R_{soft\_ACH}$, which is defined by the semantic similarity between the agent's thought process and the ACH Protocol. To quantify this, we compute the cosine similarity of their vector embeddings, generated using the BGE-M3~\cite{multi2024m3} model. This approach grants the model a more flexible exploration space.

\paragraph{Curriculum Annealing Guidance}
Simultaneously, to enable the model to explore more possibilities, we created two additional prompt types: a ``Full ACH Protocol" and a ``Simplified ACH Protocol". During training, we use a cosine annealing schedule to dynamically adjust the probability of sampling these two prompt types:
\begin{equation}
  p_{\text{full}} = \frac{1}{2} \left(1 + \cos(\pi t)\right),
\end{equation}
\begin{equation}
  p_{\text{simple}} = 1 - p_{\text{full}},
\end{equation}
where $t = \frac{i}{T}$, with $i$ denoting the current training step and $T$ the total number of training steps.

Thus, the stage two score is computed as:
\begin{equation}
  score_\text{Stage2} = score_\text{format} + score_\text{answer} + score_\text{soft\_ACH}.
\end{equation}

\section{Experiment}
To comprehensively evaluate the effectiveness, generalization capability, and robustness of our proposed AgentCDM framework, we designed a series of rigorous experiments. This section will detail the benchmark datasets, baseline methods for comparison, specific implementation details, and an in-depth analysis of the experimental results.

\subsection{Benchmarks}
To ensure a comprehensive evaluation, we selected three widely-recognized Multiple-Choice Question Answering (MCQA) benchmarks that span diverse domains and difficulty levels. These include MMLU~\cite{hendrycks2020measuring}, a large-scale benchmark covering 57 subjects to test broad knowledge; MMLU-Pro~\cite{wang2024mmlu}, a more challenging version with questions requiring complex reasoning and an expanded set of 10 options; and the most difficult subset of ARC, ARC-Challenge~\cite{clark2018think}, which contains science questions demanding multi-step reasoning. This selection allows us to robustly measure the model's decision-making and reasoning performance across a spectrum of tasks. The details of the datasets used in the experiments can be found in Appendix B.

\begin{table*}[t]
\centering

\scriptsize
\begin{adjustbox}{max width=\textwidth}
\begin{tabular}{l | c | c | cccccc |cccc}
\toprule

\multirow{2}*{\textbf{Base Model}} & \multirow{2}*{\textbf{Single Agent}} & \multicolumn{1}{c|}{\textbf{Dictatorial}} & \multicolumn{6}{c|}{\textbf{Voting-Based}} & \multicolumn{4}{c}{\textbf{ACH Protocol}} \\
& & \textbf{Informed} & \textbf{Plurality} & \textbf{Bucklin} & \textbf{Borda Count} & \textbf{IRV} & \textbf{Minimax} & \textbf{Ranked Pairs} & \textbf{Homogeneous} & \textbf{Qwen-7B-Instruct} & \textbf{Qwen-7B-R1} & \textbf{AgentCDM} \\
\bottomrule
\multicolumn{13}{l}{\textbf{MMLU}} \\
glm-4-9b-chat & 67.0 & 70.6(+3.6) & 67.7(+0.7) & 67.9(+0.9) & 67.5(+0.5) & 67.7(+0.7) & 68.0(+1.0) & 68.0(+1.0) & 74.2(+7.2) & 75.1(+8.1) & \underline{78.9(+11.9)} & \textbf{79.6(+12.6) }\\
Meta-Llama-3-8B-Instruct & 66.0 & 65.2(-0.8) & 66.5(+0.5) & 66.7(+0.7) & 66.6(+0.6) & 66.5(+0.5) & 66.7(+0.7) & 66.7(+0.7) & 68.4(+2.4) & 70.1(+4.1) & \textbf{76.8(+10.8)} & \underline{76.4(+10.4)} \\
Mistral-7B-Instruct-v0.3 & 59.4 & 50.9(-8.5) & 59.9(+0.5) & 59.9(+0.5) & 61.0(+1.6) & 60.9(+1.5) & 60.9(+1.5) & 60.9(+1.5) & 61.2(+1.8) & 64.0(+4.6) & \underline{72.7(+13.3)} & \textbf{74.3(+14.9)} \\
Qwen-2-72b & 68.7 & 72.1(+3.4) & 68.6(-0.1) & 68.6(-0.1) & 68.5(-0.2) & 68.6(-0.1) & 68.6(-0.1) & 68.6(-0.1) & 76.2(+7.5) & 74.8(+6.1) & \underline{77.2(+8.5)} & \textbf{78.1(+9.4)} \\
GPT-4 & 84.7 & \textbf{85.7(+1.0)} & 84.5(-0.2) & 84.6(-0.1) & 84.4(-0.3) & 84.5(-0.2) & 84.6(-0.1) & 84.6(-0.1) & 83.8(-0.9) & 74.7(-10) & 84.5(-0.2) & \underline{85.6(+0.9)} \\
Average & 69.2 & 69.8(+0.6) & 68.9(-0.3) & 69.4(+0.2) & 69.5(+0.3) & 69.6(+0.4) & 69.6(+0.4) & 69.8(+0.6) & 72.8(+3.6) & 71.7(+2.5) & \underline{78.0(+8.8)} & \textbf{78.8(+9.6)} \\[1pt]
\hline
\multicolumn{13}{l}{\textbf{MMLU\_PRO}} \\
glm-4-9b-chat & 47.4 & 48.3(+0.9) & 47.5(+0.1) & 47.7(+0.3) & 47.4(+0.0) & 47.5(+0.1) & 47.9(+0.5) & 47.7(+0.3) & 48.8(+1.4) & 49.5(+2.1) & \underline{57.5(+10.1)} & \textbf{63.9(+16.5)} \\
Meta-Llama-3-8B-Instruct & 41.2 & 40.8(-0.4) & 41.4(+0.2) & 41.2(+0.0) & 41.6(+0.4) & 41.4(+0.2) & 41.6(+0.4) & 41.4(+0.2) & 42.3(+1.1) & 44.5(+3.3) & \underline{54.4(+13.2)} & \textbf{63.7(+22.5)} \\
Mistral-7B-Instruct-v0.3 & 32.5 & 31.7(-0.8) & 32.0(+0.5) & 32.0(+0.0) & 31.9(-0.6) & 32.0(-0.5) & 32.1(-0.4) & 32.0(-0.5) & 31.5(-1.0) & 37.2(+4.7) & \underline{48.6(+16.1)} & \textbf{61.8(+29.3)} \\
Qwen-2-72b & 49.2 & 51.1(+1.9) & 49.0(-0.2) & 49.2(+0.0) & 49.5(+0.3) & 49.0(-0.2) & 49.3(+0.1) & 49.3(+0.1) & 53.4(+5.8) & 53.1(+5.5) & \underline{60.8(+11.6)} & \textbf{65.7(+16.5)} \\
GPT-4 & 69.0 & \textbf{71.2(+2.2)} & 69.8(+0.8) & 69.6(+0.6) & 69.4(+0.4) & 69.8(+0.8) & 69.6(+0.6) & 69.5(+0.5) & 64.2(-4.8) & 58.1(-10.9) & 68.5(-0.5) & \underline{71.0(+2.0)} \\
Average & 47.9 & 48.0(+0.1) & 48.6(+0.7) & 47.9(+0.0) & 47.9(+0.0) & 48.0(+0.1) & 48.0(+0.1) & 48.1(+0.2) & 48.0(+0.1) & 48.5(+0.6) & \underline{58.0(+10.1)} & \textbf{65.2(+17.3)} \\[1pt]
\hline
\multicolumn{13}{l}{\textbf{ARC-Challenge}} \\
glm-4-9b-chat & 84.6 & 88.8(+4.2) & 85.0(+0.4) & 85.0(+0.4) & 85.0(+0.4) & 85.0(+0.4) & 84.0(-0.6) & 84.0(-0.6) & 89.4(+4.8) & 90.8(+6.2) & \textbf{92.6(+8.0)} & \underline{91.8(7.2)} \\
Meta-Llama-3-8B-Instruct & 77.2 & 72.4(-4.8) & 77.0(-0.2) & 77.0(-0.2) & 77.0(-0.2) & 77.0(-0.2) & 77.0(-0.2) & 77.0(-0.2) & 77.4(+2.0)& 76.6(-0.6) & \textbf{81.8(+4.6)} & \underline{81.0(+3.8)} \\
Mistral-7B-Instruct-v0.3 & 63.2 & 61.2(-2.0) & 62.0(-1.2) & 62.0(-1.2) & 63.0(-0.2) & 62.0(-1.2) & 62.0(-1.2) & 62.0(-1.2) & 66.8(+3.6) & 70.2(+7.0) & \underline{81.6(+18.4)} & \textbf{84.0(+20.8)} \\
Qwen-2-72b & 86.7 & 85.6(-1.1) & 85.0(-1.7) & 85.0(-1.7) & 86.0(-0.7) & 85.0(-1.7) & 85.0(-1.7) & 85.0(-1.7) & 88.8(+2.1) & 87.6(+0.9) & \underline{90.0(+3.3)} & \textbf{91.6(+4.9)} \\
GPT-4 & 91.3 & 91.5(+0.2) & \underline{92.9(+1.6)} & 91.9(+0.6) & 91.9(+0.6) & \underline{92.9(+1.6)} & 91.9(+0.6) & 92.0(+0.7) & 92.0(+0.7) & 86.0(-5.3) & 92.0(+0.7) & \textbf{93.4(+2.1)} \\
Average & 80.6 & 80.0(-0.6) & 79.9(-0.7) & 80.4(-0.2) & 80.2(-0.4) & 80.6(+0.0) & 80.4(-0.2) & 80.0(-0.6) & 82.9(+2.3) & 82.2(+1.6) & \underline{87.6(+7.0)} & \textbf{88.4(+7.8)} \\[1pt]
\hline
All Average & 65.9 & 66.0(+0.1) & 65.8(-0.1) & 65.9(+0.0) & 65.9(+0.0) & 66.0(+0.1) & 66.0(+0.1) & 66.0(+0.1) & 67.9(+2.0) & 67.5(+1.6) & \underline{74.5(+8.6)} & \textbf{77.5(+11.6)} \\[1pt]
\bottomrule

\end{tabular}
\end{adjustbox}

\caption{Comprehensive performance comparison of AgentCDM against various baseline methods on the MMLU, MMLU\_PRO, and ARC-Challenge datasets. All values represent accuracy (\%). Numbers in parentheses denote the performance change relative to the ``Single Agent" baseline for each row. Baselines include the model's standalone performance (Single Agent), six Voting-Based methods, and an unstructured Informed Dictatorial method. The ACH Protocol group evaluates different decision agents: in the Homogeneous setting, the base model corresponding to each row is used as the decision agent, whereas Qwen-7B-Instruct and Qwen-7B-R1 serve as fixed external decision agents. For each row, the best-performing method is highlighted in bold, and the second-best is underlined. The results show that AgentCDM consistently outperforms other approaches.}
\label{table1}
\end{table*}

\subsection{Baselines}
To evaluate the effectiveness of AgentCDM, we compare it against two major categories of baseline methods: dictatorial and voting-based strategies. We adopt the baseline configurations from GEDI~\cite{zhao2024electoral}, which defines representative and standardized setups for both paradigms. This alignment ensures a fair and consistent comparison across methods.

\paragraph{Dictatorial Methods}
We evaluate an ``Informed Dictatorial" method. In this approach, multiple ``Executor Agents" first independently process the query to generate their own answers along with supporting analyses. Subsequently, a single ``Decider Agent" evaluates all of these outputs and is solely responsible for producing the system's final decision.
Crucially, this ``Decider Agent" is typically activated with an unstructured prompt that simply asks for a final judgment without providing a systematic evaluation framework, leaving the quality of the outcome entirely dependent on the agent's unguided reasoning capabilities.

\paragraph{Voting-based Methods}
Consistent with GEDI, we selected a variety of voting-based methods to aggregate agent preferences. Our selection spans several mechanism types. We employ simple systems such as Plurality, which considers only first-choice votes, and the cardinal-based Range Voting. Our evaluation also includes several ordinal systems that leverage ranked preferences: Borda Count~\cite{emerson2013original,davies2014complexity}, which assigns points based on rank; Bucklin Voting~\cite{erdelyi2015control}, which accumulates ranked choices to find a majority; and Instant-Runoff Voting (IRV)~\cite{freeman2014axiomatic}, which operates through sequential elimination. Finally, we incorporate methods based on pairwise comparisons, namely Minimax~\cite{brams2007minimax}, which seeks to minimize the largest margin of defeat, and Ranked Pairs, which constructs a final ranking from the strongest pairwise victories.

\subsection{Models and Implementation Details}
To simulate a heterogeneous agent environment, we instantiated agents using representative LLMs, including open-source models such as GLM-4-9B-Chat, Mistral-7B, LLaMA-3-8B, and Qwen-2-72B, along with the closed-source GPT-4~\cite{glm2024chatglm, jiang2023mistral7b, dubey2024llama, team2024qwen2, achiam2023gpt}. Our AgentCDM model is trained from Qwen-7B-Base~\cite{bai2023qwen}. Models are trained on A800 GPUs for 50–100 rollout–update iterations.
Each batch samples P=256 prompts, with N=5 rollouts per prompt. Policy updates use GRPO, Adam optimizer. The sampling configurations are unified (temperature = 0.6, top-p = 0.95), and n=5 responses are generated per query. For fair comparison, we also evaluate Qwen-7B-Instruct and the reasoning-focused Qwen-7B-R1 under the ACH Protocol~\cite{bai2023qwen, guo2025deepseek}. Average Accuracy is used as the primary evaluation metric. Voting-based baselines follow a 5-shot in-context learning (ICL) setup, while Informed Dictatorial and ACH Protocol operate directly on Execution Agent outputs. Additional details are provided in Appendices A and C.

\subsection{Main Results}

Our comprehensive evaluation on the MMLU, MMLU\_PRO, and ARC-Challenge benchmarks (Table~\ref{table1}) demonstrates the clear advantages of the proposed AgentCDM framework. Trained using our two-stage paradigm, AgentCDM consistently and significantly outperforms all baselines—including individual models, the unstructured Informed Dictatorial approach, various voting-based strategies, and models solely guided by the ACH protocol. The gains are especially pronounced on challenging tasks. For instance, on MMLU-PRO, AgentCDM improves the performance of Meta-Llama-3 and Mistral-7B by 22.5 and 29.3 percentage points over the Single Agent baseline, respectively, highlighting its ability to facilitate effective collaborative decision-making.

Table~\ref{table1} also reports results for Qwen-7B-Instruct and Qwen-7B-R1 under ACH protocol guidance. Notably, Qwen-7B-R1 is a reasoning-optimized model with the same parameter scale. Yet, our AgentCDM, trained from a base model, achieves higher average accuracy when guided by the same protocol. For example, on the challenging MMLU-PRO benchmark, AgentCDM achieves 65.2\% accuracy, outperforming Qwen-7B-R1’s already strong 58.0\%. This result highlights the effectiveness of our two-stage training framework in enhancing structured reasoning, even compared to specialized reasoning models.

Further analysis reveals the inherent power of structured reasoning. Remarkably, even without parameter tuning, introducing the ACH Protocol as a zero-shot prompting strategy significantly enhances performance. For example, the Homogeneous variant—which functions as both executor and decision-maker—generally outperforms the unstructured Informed Dictatorial method when guided by the ACH Protocol, surpassing it by 1.9 percentage points on the All Average metric and consistently outperforming all Voting-Based methods. These results show that a systematic, evidence-driven evaluation of competing hypotheses is, in itself, an effective approach to promoting rigorous, bias-resistant reasoning.

\subsection{Cross-Dataset Generalization Evaluation}

\begin{table}[t]
\centering
\begin{adjustbox}{max width={.95\columnwidth}}
\begin{tabular}{l c c c}
\toprule
 & \multicolumn{3}{c}{\textbf{Testing Dataset}} \\ 
 \cmidrule(lr){2-4}
\textbf{Training Dataset} & \textbf{MMLU} & \textbf{MMLU\_PRO} & \textbf{ARC-Challenge} \\
\midrule
MMLU & \textbf{78.5} & 55.0 & 91.0 \\
MMLU\_PRO & 80.8 & \textbf{65.2} & 94.0 \\
ARC-Challenge & 74.9 & 52.0 & \textbf{89.5} \\
\bottomrule
\end{tabular}
\end{adjustbox}
\caption{Cross-dataset generalization performance of the AgentCDM framework. Each row specifies the training dataset, and each column specifies the evaluation dataset. Values indicate accuracy(\%). Diagonal results (in bold) show in-domain performance.}
\label{table2}
\end{table}

To further assess the generalization capability of our AgentCDM framework, we conducted cross-dataset transfer evaluations. Specifically, a decision-making agent trained on a particular dataset (e.g., MMLU-PRO) was directly evaluated on previously unseen test sets (e.g., MMLU and ARC-Challenge).

As shown in the table~\ref{table2}, models trained on the more challenging MMLU-PRO dataset exhibited remarkable generalization performance. Notably, the model achieved a score of 80.84 on the MMLU test set and 94.0 on the ARC-Challenge test set. These results not only demonstrate strong performance but even surpass the scores of models trained directly on MMLU and ARC-Challenge (which achieved 78.52 and 89.5, respectively).

These findings provide strong evidence that training on more complex and demanding datasets enables AgentCDM to acquire deeper, more generalizable structured reasoning capabilities. Such capabilities transcend specific knowledge domains and can be effectively transferred to diverse tasks, leading to high-quality decision-making across domains. In contrast, models trained on relatively simpler datasets (e.g., ARC-Challenge) perform poorly when transferred to more challenging tasks (e.g., MMLU-PRO), further highlighting the critical role of high-quality and high-difficulty training data in cultivating robust general decision-making abilities.

\subsection{Robustness and Scalability Analysis}

\begin{figure}
    \centering
    \includegraphics[width=1\linewidth,height=4.5cm]{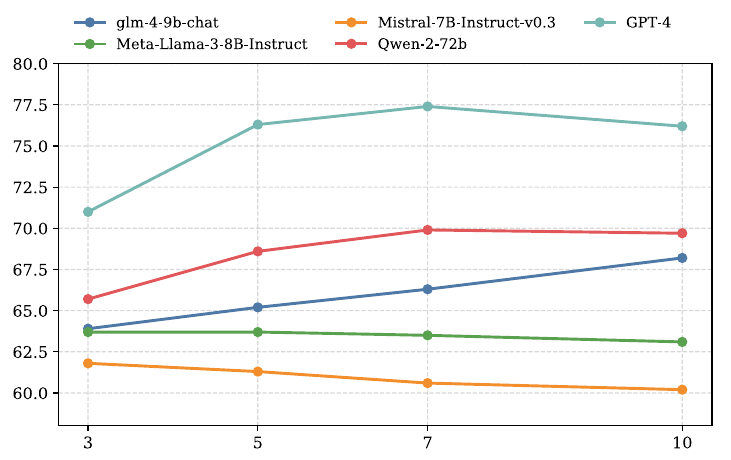}
    \caption{Effect of agent quantity on MMLU-PRO performance. Values indicate accuracy (\%).}
    \label{fig4}
\end{figure}

\begin{table}[t]
\centering
\begin{adjustbox}{max width={.95\columnwidth}}
\begin{tabular}{l c c}
\toprule
\textbf{Dataset} & \textbf{Qwen-7B-Instruct(ACH)} & \textbf{AgentCDM} \\
\toprule
MMLU      & 75.2 & \textbf{78.9} \\
MMLU\_PRO & 55.9 & \textbf{67.9} \\
ARC-Challenge       & 87.2 & \textbf{93.0} \\
\bottomrule
\end{tabular}
\end{adjustbox}
\caption{Performance evaluation in a heterogeneous agent pool. Values indicate accuracy (\%).}
\label{table4}
\end{table}

To evaluate AgentCDM's performance in settings that resemble real-world applications, we conducted experiments on its scalability and robustness.

First, we tested scalability by varying the number of agents on the MMLU-PRO dataset (Figure~\ref{fig4}), uncovering a stark dual phenomenon. For capable models (e.g., GPT-4), performance scaled positively with more agents, demonstrating effective ``collective intelligence". Conversely, for less capable models (e.g., Mistral-7B), performance degraded as the number of agents increased. We conclude that this is because aggregating multiple weaker inputs amplifies noise more than signal, overwhelming the decision-making process.

Second, to assess the robustness of decision-makers in more realistic settings, we build a heterogeneous agent pool comprising outputs from five distinct models. For each query, the decision-making agent receives as input a combination of three outputs randomly selected from this pool. As shown in Table~\ref{table4}, the fully trained AgentCDM significantly outperformed Qwen-7B-Instruct with ACH Protocol, achieving a score of 67.9 versus the baseline's 55.9 on MMLU-PRO. This result, combined with the scalability findings, strongly validates our two-stage training paradigm. It successfully endows the agent with internalized analytical skills to critically evaluate diverse and conflicting information, demonstrating the robustness crucial for complex, dynamic environments.

\subsection{Ablation Study}

\begin{table}[t]
\centering
\begin{adjustbox}{max width={.95\columnwidth}}
\begin{tabular}{l c c c}
\toprule
\textbf{Method} & \textbf{MMLU} & \textbf{MMLU\_PRO} & \textbf{ARC-Challenge} \\

\toprule
    Full & \textbf{78.5} & \textbf{65.2}  & \textbf{96.0}  \\
    Vanilla prompt & 72.6  & 52.5 & 89.0  \\
    Scaffolding-Only & 68.1 & 49.4  & 84.0  \\
    Exploration-Only & 71.5 & 52.1 & 89.7  \\
\bottomrule
\end{tabular}
\end{adjustbox}
\caption{Ablation study results on three benchmarks. ``Full" is our complete two-stage AgentCDM framework. ``Scaffolding-Only" and ``Exploration-Only" refer to using only Stage 1 or Stage 2 of our training, respectively. Values indicate accuracy (\%).}
\label{table5}
\end{table}

To validate the necessity of our proposed two-stage training paradigm, we conducted a series of ablation studies (Table~\ref{table5}) to isolate the contributions of Stage 1 (Structured Scaffolding) and Stage 2 (Autonomous Exploration). The results clearly reveal their synergistic effect. The Full model (AgentCDM), incorporating both stages, consistently achieved the best performance across all datasets. In contrast, removing the second stage (Scaffolding-Only) led to a drastic performance drop, indicating that rigid adherence to the structured protocol impairs generalization. Conversely, removing the first stage (Exploration-Only) resulted in mediocre performance comparable to a simple baseline, demonstrating that unguided exploration is inefficient without a solid foundation in structured reasoning. These experiments provide compelling evidence that the two stages are both indispensable and complementary: Stage 1 endows the model with a robust reasoning core, while Stage 2 fosters the generalization and adaptability crucial for superior performance.

\section{Conclusion and Limitations}
In this work, we introduced AgentCDM, a novel framework for collaborative decision-making in LLM-based multi-agent systems, inspired by the ACH from cognitive science. Through a two-stage training process---combining structured reasoning scaffolding and staged autonomous exploration---we systematically mitigated cognitive bias and enabled agents to arrive at more accurate, robust, and generalizable decisions. Extensive experiments across diverse benchmarks validated the effectiveness of AgentCDM, showcasing consistent improvements over both voting-based and dictatorial baselines, particularly in challenging and heterogeneous settings.

However, our approach is not without limitations. The effectiveness of AgentCDM is influenced by the quality and diversity of the agent-generated hypotheses from the execution phase. Additionally, the current framework is primarily designed for cooperative agents; its performance in adversarial or noisy agent environments remains an open question for future exploration. Future work will investigate more adaptive scaffolding strategies, methods to improve robustness against unreliable agents, and applications in broader real-world decision-making contexts.

\newpage
\bibliography{agent_cdm}

\end{document}